# GLStyleNet: Higher Quality Style Transfer Combining Global and Local Pyramid Features

Zhizhong Wang* , Lei Zhao* , Wei Xing , Dongming Lu

College of Computer Science and Technology, Zhejiang University

{endywon, cszhl, wxing, ldm}@zju.edu.cn

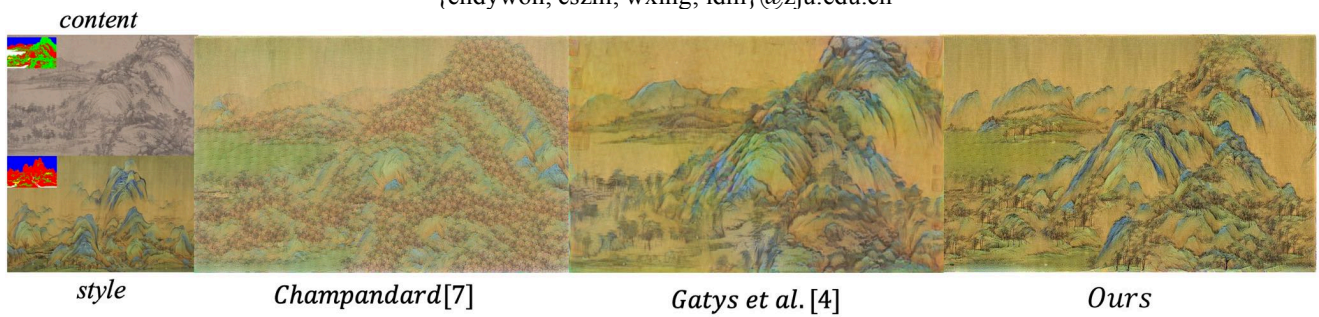

Figure 1: Overall style transfer of Chinese ancient paintings. Neither [7] (column 2) nor [4] (column 3) can get satisfying results. Our method results the most exquisite painting work (last column). The content image is named "Dwelling in the Fuchun Mountains" by Gongwang Huang in Yuan Dynasty (column 1, row 1), and the style image is named "A Thousand Li of Rivers and Mountains" by Ximeng Wang in Song Dynasty (column 1, row 2).

## Abstract

*Recent studies using deep neural networks have shown remarkable success in style transfer especially for artistic and photo-realistic images. However, the approaches using global feature correlations fail to capture small, intricate textures and maintain correct texture scales of the artworks, and the approaches based on local patches are defective on global effect. In this paper, we present a novel feature pyramid fusion neural network, dubbed GLStyleNet, which sufficiently takes into consideration multi-scale and multi-level pyramid features by best aggregating layers across a VGG network, and performs style transfer hierarchically with multiple losses of different scales. Our proposed method retains high-frequency pixel information and low frequency construct information of images from two aspects: loss function constraint and feature fusion. Our approach is not only flexible to adjust the trade-off between content and style, but also controllable between global and local. Compared to state-of-the-art methods, our method can transfer not just large-scale, obvious style cues but also subtle, exquisite ones, and dramatically improves the quality of style transfer. We demonstrate the effectiveness of our approach on portrait style transfer, artistic style transfer, photo-realistic style transfer and Chinese ancient painting style transfer tasks. Experimental results indicate that our unified approach improves image style transfer quality over previous state-of-the-art methods, while also accelerating the whole process in a certain extent. Our code is available at https://github.com/EndyWon/GLStyleNet.*

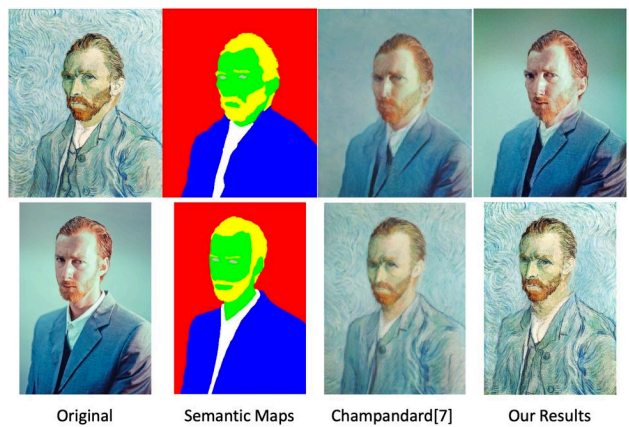

Figure 2: Comparison of portrait style transfer. Note that the results produced by Champandard[7] (column 3) are blurred and some details are not plausible enough (see noses, eyes, ears, clothes and backgrounds), while our results are more faithful and clear (last column).

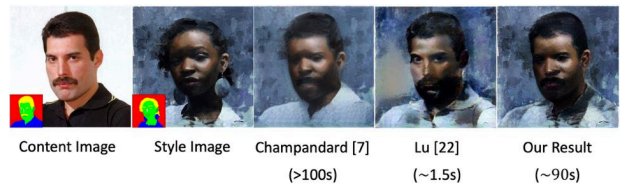

Figure 3: Another comparison of portrait style transfer. Note that although the speed of Lu's method[22] is fastest, the result is not expressive (column 4). Our method produces the most faithful result (see ears, mouths and backgrounds) and consumes less time than [7] while integrating global constraint (see last column and column 3).

*These Authors equally contributed to the work.

# 1. Introduction

Style transfer, or to repaint an existing photograph with the style of another, is considered a challenging but interesting problem in arts[1]. Recently, this task has become an active topic both in academia and industry due to the influential work by Gatys et al.[2], where the Gram matrices based on the feature maps are extracted from deep neural networks to represent global feature correlations about the style image, and achieves visually stunning results. Unfortunately, some local pixel details are lost in image style transfer process, the method based on Gram matrices is more suitable for whole artistic style transfer, especially with painterly, sketch or abstract styles. To remedy this, Li and Wand[6] combine markov random fields (MRFs) and convolutional neural networks to replace Gram matrices, which maintains local patterns of the style image. To further improve the transfer results between the corresponding regions of the style and content images, Champandard[7] extends [6] to incorporate the semantic maps into patch-matching process. Though past work creates visually pleasing results based on different types of style expression, two important drawbacks stand out when we apply these method on Chinese ancient painting style transfer:

(1) Existing methods cannot retain both structural information and microscopic high-frequency information. The approaches using global feature correlations fail to capture small, intricate textures and maintain correct texture scales of the artworks, and the approaches based on local patches are defective on global effect.

(2) The produced results are blurred and some transferred details are not plausible enough (e.g. Figure 2), especially on the Chinese ancient painting style transfer (e.g. Figure 1).

Inspired by the global gram-based method[4], the local semantic patch-based method[7] and Holistically-Nested Edge Detection[8], we propose a novel deep convolutional neural network architecture (GLStyleNet) for high quality image style transfer. Our contribution is fourfold:

(1) We present a novel forward deep neural network model dubbed GLStyleNet, which best aggregates layers across a VGG network. The outputs from GLStyleNet are multi-scale and multi-level, with the side-output-plane size becoming smaller and the receptive field size becoming larger. One aggregation process is proposed to automatically combine the multi-scale outputs from multiple layers.

(2) Our method also firstly combines the global and local attributes of images by global loss function constraint based on Gram matrices and local loss function constraints based on patch-matching of high level image features. Our method retains high-frequency pixel information and low frequency construct information of images and dramatically improves the quality of style transfer.

(3) Our approach is not only flexible to adjust the trade-off between content and style, but also controllable between global and local.

(4) Our approach speeds up the whole process with the least texture losing in a certain extent compared to Champandard[7] when improves image style transfer quality (e.g. Figure 3).

# 2. Related Work

**Gram-based Style Transfer:** Before Gatys et al. first use deep neural networks[9,10] in both texture synthesis[3] and style transfer[4] (which named neural algorithm of artistic style[2,5]), researchers tried to solve these problems by matching the statistics of content image and style image[11,12]. However, compared with the traditional methods only consider low-level features during the process, deep neural networks can extract not only low-level features but also high-level features, this results the new images generated by combining high-level information from content image with multi-level information from style image are usually more impressive.

The method proposed by Gatys et al. is gram-based, using Gram matrices to represent style features, which is computationally expensive due to the iterative optimization procedure. On the contrary, many deep learning methods use the perceptual objective computed from a neural network as the loss function to construct feed-forward neural networks to synthesize images[13,14,15]. Fast style transfer has achieved great results and is receiving a lot of attentions. Johnson et al.[16] propose a feed-forward network for both fast style transfer and super resolution using the perceptual losses defined in Gatys et al.[4]. A similar architecture texture net is introduced to synthesize textured and stylized images[17]. More recently, Ulyanov et al.[18] show that replacing spatial batch normalization [19] in the feed-forward network with instance normalization can significantly improve the quality of generated images for fast style transfer. Here the method presented by this paper is based on optimization procedure, so the speed is slower than the methods [16] and [17].

**Patch-based Style Transfer**: while gram-based method can get astonishing results especially in artistic style transfer, it only considers the global correlations but ignores the local pixel details. An alternative method proposed by Li and Wand[6] combines markov random fields (MRFs) and convolutional neural networks to replace Gram matrices, which is patch-based, using a nearest neighbor calculation to match the 3×3 neural patches extracted from outputs of convolutional layers with the input of style and content images. Based on this method, Champandard[7] incorporates the semantic maps into patch-matching process and improves the effect of semantic style transfer, which aims to transfer the style from the regions of style image to the corresponding semantically similar regions of content image. Instead of combining MRF prior, Chen and Hsu[20] provide an alternative approach to exploit masking out process to constrain the spatial correspondence and a higher order style feature statistic to further improve the results[21]. Later, Lu et al.[22] address the computation bottleneck of [7] and propose a fast semantic style transfer method, which optimize the objective function in feature space, instead of in image space, by a pre-trained decoder network, using a lightweight feature reconstruction algorithm. More recently, Mechrez et al.[23] propose an alternative contextual loss to realize semantic style transfer in a segmentation-free manner[21].

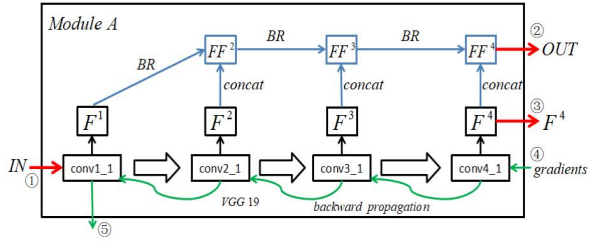

Figure 4: Module A: The deep convolutional network module of GLStyleNet. The feature map extracted from layer {conv$i$\_1} of a VGG19 network is denoted as $F^i$. As the blue arrows and rectangles show, the aggregation begins at the shallowest, largest resolution and then iteratively fuses deeper, smaller resolutions. We use bilinear resizing (BR) to downsample the intermediate feature maps, and concatenate (concat) operations to aggregate feature maps from different layers. The process of gradient backward propagation is shown in green flows.

## 3. Global and Local Style Network

GLStyleNet is a unified model consists of four modules, a deep convolutional network module that best aggregates layers across a VGG network and three other modules facilitate the combination of global and local features.

### 3.1. Layer Aggregation

As the blue arrows and rectangles shown in Figure 4, we make a layer aggregation based on the outputs of the first 4 convolutional layers {conv1\_1, conv2\_1, conv3\_1, conv4\_1} from a VGG19 network[9]. Since the shallow layers extract spatially finer but less semantic features and deeper layers obtain more semantic but spatially coarser features, we aggregate iteratively from the shallowest layer to deeper layers so that to fuse the semantic and spatially fine features simultaneously. But the output resolution of the shallow layer is larger than that of the deeper layer because of the multiple average pooling operations (we replace the maximum pooling in VGG19 with the average pooling). So before we aggregate these output features of different layers, we should firstly resize them to the same size with the main characteristics remain unchanged, here we use bilinear resizing (BR) and concatenate (concat)

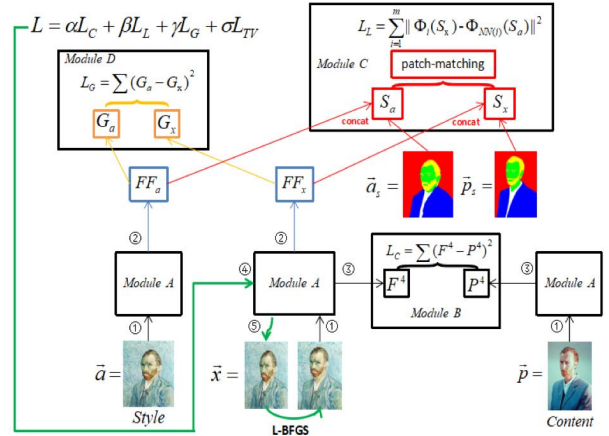

Figure 5: Overview of our proposed model, GLStyleNet, which consists of four modules. Module A: a deep convolutional network module best aggregates layers across a VGG network as shown in Figure 4. Module B: a content module computes the content loss (right, bottom, in black). Module C: a local module first incorporates semantic annotations, then extracts and matches local patches of fused style features, and finally computes the local loss (right, top, in red). Module D: a global module computes the global correlations and global loss of fused style features (left, top, in yellow). Our total loss $L$ (left, top) contains a content loss $L_C$, a local style loss $L_L$, a global style loss $L_G$ and a total variation regularization loss $L_{TV}$, we use L-BFGS to optimize the total loss (middle, in green).

operations to iteratively aggregate adjacent layers across a VGG19 network.

As shown in Figure 4, Let $F^i$ denotes the output feature map of layer {conv$i$\_1}, $BR(C, D)$ denotes the bilinear resizing that downsample $C$ to the same height and width as $D$ with channel remains unchanged, $\oplus$ denotes the concatenate operation on the channel axis, $FF^i$ denotes that we iteratively aggregate the feature maps from the layer {conv1\_1} to the {conv$i$\_1}. Our layer aggregation process is defined as follows:

$$FF^i = \begin{cases} BR(F^1, F^2) \oplus F^2 & i = 2 \\ BR(FF^{i-1}, F^i) \oplus F^i & i > 2 \end{cases} \quad (1)$$

### 3.2. Combining Global and Local Pyramid Features

Our proposed model consists of four modules, as shown in Figure 5, jointly facilitate global and local combination. let $\vec{p}$ and $\vec{x}$ be the content image and the image that is generated ($\vec{x}$ can be initialized from content image, style image or white noise, the effect of different initialization will be discussed in Section 4.1), and $P^4$ and $F^4$ denote the feature representation in layer {conv4\_1}, respectively. In module B, the content loss is defined as the square-error loss between this two feature representations:

$$L_C(\vec{x}, \vec{p}) = \sum_{i,j} (F^4_{ij} - P^4_{ij})^2 \quad (2)$$

Let $\vec{a}$ denotes the style image, $\vec{a}_s$ and $\vec{p}_s$ denote the style semantic map and content semantic map, respectively. The style loss contains two parts, a local style loss $L_L$ and a global style loss $L_G$, combines the local and global style features. These two losses are all computed on the pyramid fused features (see section 3.1) outputted from module A, the fused features of style image and generated image are denoted as $FF_a$ and $FF_x$, respectively. Then for Module C, we first concatenate the fused features with the semantic maps on the channel axis (we give a parameter $\beta_1$ to weight the semantic map channels), which are denoted as $S_a$ and $S_x$, respectively. And then for $S_a$ and $S_x$, we respectively extract the local patches of the size 3×3×d, here the $d$ is the number of channels for the concatenated features. The lists of all patches of $S_a$ and $S_x$ are denoted as $\Phi(S_a)$ and $\Phi(S_x)$, for each patch $\Phi_i(S_x)$, we find the best matching patch $\Phi_{NN(i)}(S_a)$ over all $m_a$ patches in $\Phi(S_a)$, here $m_a$ is the cardinality of $\Phi(S_a)$, based on normalized cross-correlation measure:

$$NN(i) := \underset{j=1,\ldots,m_a}{\mathrm{argmax}} \frac{<\Phi_i(S_x), \Phi_j(S_a)>}{\|\Phi_i(S_x)\| \cdot \|\Phi_j(S_a)\|} \quad (3)$$

Finally, our local style loss $L_L$ is defined as follows:

$$L_L(S_x, S_a) = \sum_{i=1}^{m} \|\Phi_i(S_x) - \Phi_{NN(i)}(S_a)\|^2 \quad (4)$$

Here $m$ is the cardinality of $\Phi(S_x)$.

For Module D, we first compute the global correlations on the fused features $FF_a$ and $FF_x$. Take $FF_a$ as an example, suppose the size of $FF_a$ is h×w×c, here h, w, c respectively denotes the height, width and channel of fused feature map $FF_a$, we first reshape the $FF_a$ as the size of N×M, where N=c and M=h×w, then compute the global correlation matrix:

$$G_a = <FF_a, FF_a^T> \quad (5)$$

We denote the global correlation matrices computed on $FF_a$ and $FF_x$ as $G_a$ and $G_x$, respectively, and then our global style loss $L_G$ is defined as follows:

$$L_G(G_a, G_x) = \frac{1}{4N^2M^2} \sum_{i,j} (G_{aij} - G_{xij})^2 \quad (6)$$

In the last, for the generated image $I$ with the height $H$, width $W$ and channels $C$, we use total variation regularization loss $L_{TV}$ to obtain spatially smoother results:

$$L_{TV}(I) = \sum_{i=1}^{H-1}\sum_{j=1}^{W}\sum_{k=1}^{C}(I_{i+1,j,k} - I_{i,j,k})^2 + \sum_{i=1}^{H}\sum_{j=1}^{W-1}\sum_{k=1}^{C}(I_{i,j+1,k} - I_{i,j,k})^2 \quad (7)$$

In conclusion, our total loss $L$ contains a content loss $L_C$ (weighted by α), a local style loss $L_L$ (weighted by β), a global style loss $L_G$ (weighted by γ) and a total variation regularization loss $L_{TV}$ (weighted by σ). The total loss $L$ is defined as follows:

$$L = \alpha L_C + \beta L_L + \gamma L_G + \sigma L_{TV} \quad (8)$$

We use L-BFGS to optimize the total loss as shown in green flows in Figure 5.

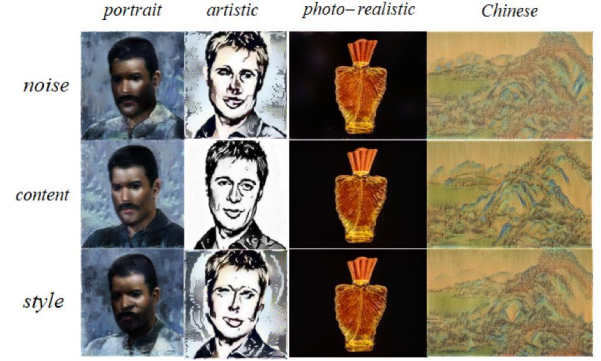

Figure 6: The results of different initialization, each pair of samples will be demonstrated in the section 4. Notice that for portrait style transfer, initializing from style image will get more expressive result (column 1), while for artistic (column 2) and Chinese ancient painting (column 4) style transfer, it is more reasonable to initialize from content image. Last for photo-realistic style transfer (column 3), results obtained from different initialization are similar, but it can preserve more content details through initializing from content image.

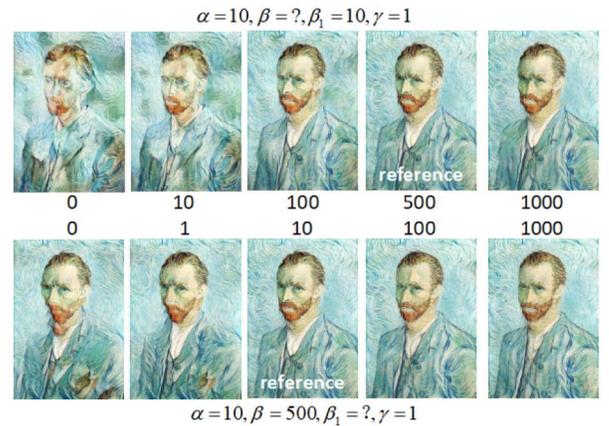

Figure 7: Varying parameters $\beta$ and $\beta_1$ for style transfer, the generated images are all initialized from style image. First row shows changes of local style weight $\beta$, note that as $\beta$ increases, more style detail information is maintained, but glitches occur when $\beta$ is too high. Second row shows changes of semantic weight $\beta_1$, less semantic constraint as $\beta_1$ decreases, but too high $\beta_1$ lowers quality of style.

## 4. Experimental Results

we first discuss the initialization for L-BFGS, explain the trade-off between parameters, and then demonstrate the effect of layer aggregation and global and local combination to understand how the GLStyleNet contribute. We evaluate the effectiveness of our approach by showing the effect on four style transfer tasks (portrait style transfer, artistic style transfer, photo-realistic style transfer and Chinese ancient painting style transfer).

### 4.1. Initialization and Parameters

As show in Figure 5 (middle), our optimization process is implemented by using L-BFGS, which the generated results will be affected by initialization. Be different from Champandard[7] only uses content image or white noise to initialize, we also use the resized style image (the style image should firstly be resized to the same size as content image). Figure 6 shows the results of different initialization. Generally, for images with similar structures, e.g. portraits, initializing from style image will get more expressive results (see column 1), while if we want to preserve more content information, it's more reasonable to initialize from content image (see column 2 to 4).

For parameters, generally, the weight α of content loss $L_C$ is fixed to 10 and the weight $\sigma$ of total variation regularization loss $L_{TV}$ is fixed to 1, Figure 7 shows a grid with visualizations of results as $\beta$ and $\beta_1$ vary, and Figure 8 shows results as $\gamma$ varies. We conclude the following:

● More style detail information is maintained as $\beta$ increases, but too high $\beta$ may cause glitches.

● As $\beta_1$ decreases, the algorithm reverts to its semantically unaware version that ignores the semantic map provided, but the quality of style will be low if the $\beta_1$ is too high. This is exactly the same as the method proposed by Champandard[7].

● The weight $\gamma$ gives a global constraint not only improves the style detail performance, but also increases the semantic awareness. It may dramatically improve the results if used carefully, and control the fineness of style transfer to get more abstract or more exquisite results, as shown in Figure 12.

### 4.2. Effect of Layer Aggregation

We compare eight different schemes to demonstrate the effect of layer aggregation, including five different ways of aggregation (see a, d, e, f, g in Figure 9) and three different ways of non-aggregation (see b, c, h in Figure 9). We set parameter $\gamma$ to zero so that to exclude the influence of global constraint. Figure 10 shows more details about each scheme, the numbers in brackets respectively indicate the height, width and channel of feature maps, and the process of layer aggregation is represented by flows. The qualitative effect of layer aggregation can be observed in Figure 9, the scheme (g) (column 3, row 2) which fuses the most features achieves the best result with the least texture losing, the scheme (h) (column 4, row 2) is used in [7] by Champandard, and scheme (b) and (c) (column 2 and 3 in row 1) are discussed in [22] by Lu et al.

On the other hand, we compare the quantitative effect of layer aggregation in Figure 11. Compared with scheme (h) used in [7] (see last row in left Table), our scheme (g) can obtain more than 25% acceleration for each iteration (see the row marked in red bold in left Table), note that this will save a lot of time as the number of iterations increases, and this is simultaneously used in computing global loss as discussed in Section 3, finally results the acceleration of whole process with the least texture losing (see Figure 2). The relationship between the amounts of pixels and mean time spent on each iteration is shown in right Line Chart, they almost maintain a positive correlation. Although scheme (c) only uses single layer {conv4_1} and consumes the least time for each iteration (see the row marked in black bold in left Table), our scheme (g) can get better result with less texture losing while only spending less than 10% time more than (c).

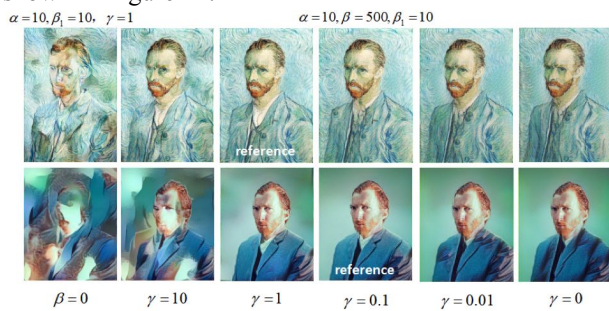

Figure 8: Varying parameter $\gamma$ to combine global and local fused features. The last column shows only local fused features without global information ($\gamma = 0$); the first column shows only global fused features without local information ($\beta = 0$); the second to fifth columns show that combining global constraint with local fused features could dramatically improve the transferred results if $\gamma$ is adjusted carefully.

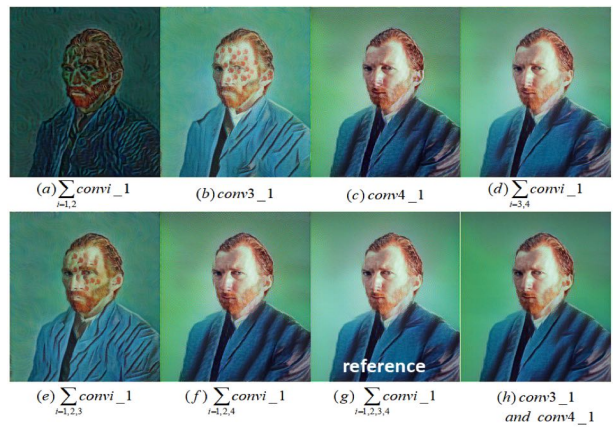

Figure 9: Comparison of qualitative effect for eight different layer aggregation schemes. See more details about each scheme in Figure 10. The scheme (g) (column 3, row 2) fuses the most features and achieves the best result, the scheme (h) (column 4, row 2) is used in [7] and the scheme (b) and scheme (c) (column 2 and column 3, row 1) are discussed in [22].

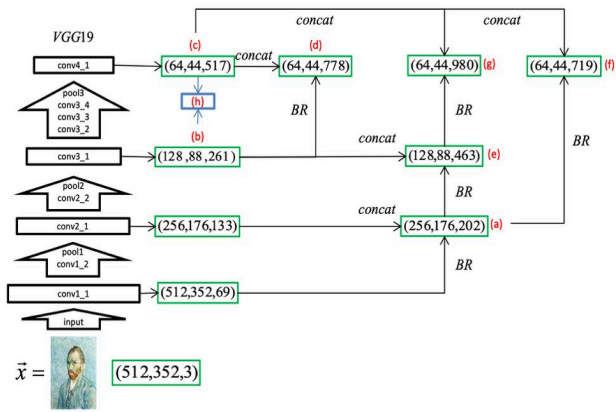

Figure 10: Details about each scheme in Figure 9, the numbers in brackets respectively indicate the height, width and channel of feature maps. We use bilinear resizing (BR) and concatenate (concat) operations to fuse the intermediate feature maps from different layers of VGG19, blue box and arrows denote using two feature maps simultaneously, i.e. scheme (h).

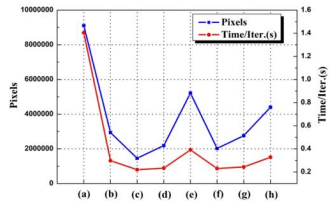

Figure 11: The amounts of pixels to calculate and mean computation time spent on each iteration of each scheme in Figure 9. Time are taken for input image of resolution 512× 352 on two NVIDIA Tesla P100 GPUs. In **left Table**, the second column represents the size of feature maps that each scheme directly operates on, the third column represents the amounts of pixels to calculate and the last column represents mean time spent on each iteration of each scheme. The least time-consuming scheme is marked by black bold, and our finally adopted scheme is marked by red bold. In **right Line Chart**, we compare the trend of amounts of pixels and mean time spent on each iteration, they almost maintain a positive correlation.

### 4.3. Effect of Combining Global and Local Pyramid Features

We have demonstrated and discussed the varieties brought by $\gamma$ in section 4.1, as shown in Figure 8, only using local fused features cannot obtain satisfying results (see last column in Figure 8), these results can be improved through appropriately integrating some global information (see column 3 and column 4 in Figure 8). For better demonstration, we carry out experiments on Chinese ancient paintings, as shown in Figure 12. Obviously, the color of the whole painting is richer when integrating more global information (see column 3 in Figure 12), and the painting would become abstract when the $\gamma$ is high (see column 4 in Figure 12), so we can control the fineness of style transfer to get more abstract or more exquisite results by adjusting parameter $\gamma$. See more details and experiments of Chinese ancient painting style transfer in section 4.7.

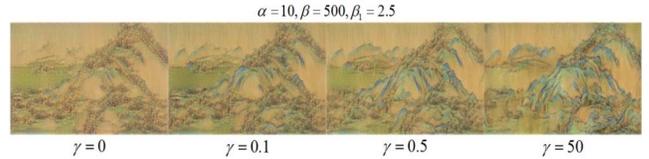

Figure 12: Effect of combining global and local fused features in Chinese ancient painting style transfer. The color information is richer while integrating more global information (increased by $\gamma$), and the whole image becomes abstract and details become missing when $\gamma$ is high (see last column). In order to make the effect more obvious, we post-process the results to increase brightness and clarity.

### 4.4. Portrait Style Transfer

Our proposed GLStyleNet performs better in semantic style transfer than state-of-the-art methods. Take portrait style transfer as an example like [7], the results produced by Champandard[7] are quite blurred and some details are not plausible enough, see Figure 2, in comparison, our results are more faithful and clear. We also compare our method with Lu et al.[22], as shown in Figure 3, although the speed of Lu's method is fastest, their result is not expressive, our method results the most expressive performance and still has speed-up compared with [7] while global constraints are added.

### 4.5. Artistic Style Transfer

We evaluate our proposed approach with four state-of-the-art methods for artistic style transfer based on the prior work of X.Li et al.[27]. As shown in Figure 13, these four sample pairs and five results from the third column to the seventh column are all from [27]. They compare their proposed method (column 7) with the optimization based approach [4] (column 3), feature transform based approaches [24] (column 4) and [25] (column 5), and the latest new approach [26] (column 6). Our results are show in the last column, all generated images are initialized from content image. Note that our proposed model is not only effective for semantic style transfer, but also for artistic style transfer without semantic annotations.

### 4.6. Photo-realistic Style Transfer

Our proposed approach is flexible to adjust the trade-off between content and style during style transfer by adjusting parameters. So it can be applied in photo-realistic style transfer, which prefers preservation of content images for both the global structures and the detailed contours during style transfer[27]. In the same way, we evaluate our proposed approach with three state-of-the-art methods based on the prior work of X.Li et.al.[27]. As shown in Figure 14, these four sample pairs and three results from the third column to the fifth column are all from [27]. They compare their proposed method (column 5) with the recent work [28,29] (column 3, column 4). Our results are show in the last column, to preserve more content information, all generated images are initialized from content image. Be different from [27] separately processes the masked regions in spatial masks and finally combines them together, our approach takes the spatial masks as the semantic maps and directly process the whole images. But this may cause some structural artifacts especially for subjects that have strict symmetry properties, e.g. the bottom image in our results, this will not happen if we process separately according to the spatial masks.

### 4.7. Chinese ancient painting style transfer

Chinese ancient painting style transfer is one of the most interesting but challenging tasks in style transfer, which is never discussed and demonstrated before. In this paper, our objective is to transfer the style of "A Thousand Li of Rivers and Mountains" by Ximeng Wang in Song Dynasty onto another Chinese ancient painting "Dwelling in the Fuchun Mountains" by Gongwang Huang in Yuan Dynasty. Please note that it is not as simple as only coloring the content image like the third column of Figure 1, the method proposed by Gatys et al.[4] get very rough results in this case. The most crucial and difficult point is how to deal with the style transfer of corresponding categories accurately under the premise of the overall effect is beautiful, e.g. transfer the style of mountains onto the mountains, the style of trees onto the trees, the style of rivers onto the rivers, and the overall combination looks still very harmonious. In other words, it is an elaborate style transfer, and may be influenced by the material of papers, the noises generated by breakages and some other irrelevant contents.

Figure 1 shows the overall style transfer of Chinese ancient paintings, we post-process all the results with the same brighten and clarity operations to make the effect more obvious. Neither local-only method proposed by Champandard[7] (see column 2) nor global-only method proposed by Gatys et al.[4] (see column 3) can get satisfying results, however, our combining global and local pyramid features method can retain local details as much as possible while also performs well on global features such as colors, which results exquisite painting work.

In order to show more details about the Chinese ancient painting style transfer, we respectively crop two local regions from the content painting and style painting, as show in Figure 15, we use more elaborate semantic map to constraint the style transfer of leaves, trunks, houses and backgrounds. The results are shown in the last column, our method can get good detail performance. This also proves that if we use our method to deal with each local area of a super-large painting, and then use appropriate method to stitch them together, we will eventually get a very exquisite work.

## 5. Conclusions

In this work, the two means are utilized to integrate the global and local statistical properties of images. Firstly, the integrated neural network, dubbed GLStyleNet, is designed to fuse pyramid features of content image and style image. As we know, the lower output features of VGG network retain much more high frequency details of image, however, the higher output features retain much more low frequency details of image. That is to say, the lower output features of VGG network retain local details (local information) of image (corresponding to larger feature scale), and the higher output features retain structure information (global information) of image (corresponding to smaller feature scale). Our GLStyleNet fuses multi-scale and multi-level pyramid features from VGG network. Secondly, our loss function constraint is based on global and local statistical attributes of images. The global statistics are based on gram matrices and the local statistics are based on feature patches. It is worth mentioning that the parameters could be learned automatically by extending and training GLStyleNet like [16], to replace complex manual parameter adjustment process. Our method dramatically improves the quality of style transfer, and the novel idea of layer aggregation helps us to accelerate the whole process with the least texture losing. Our approach is not only flexible to adjust the trade-off between content and style, but also controllable between global and local, so it can be applied in numerous style transfer tasks especially for Chinese ancient painting style transfer which is never demonstrated before. Experimental results demonstrate that our proposed approach performs better than many state-of-the-art methods, and this could assist us to create more elaborate works. In future, we will explore our method in high fidelity image style transfer and generate exquisite details of transferred image while global structure accord with constraint of content image.

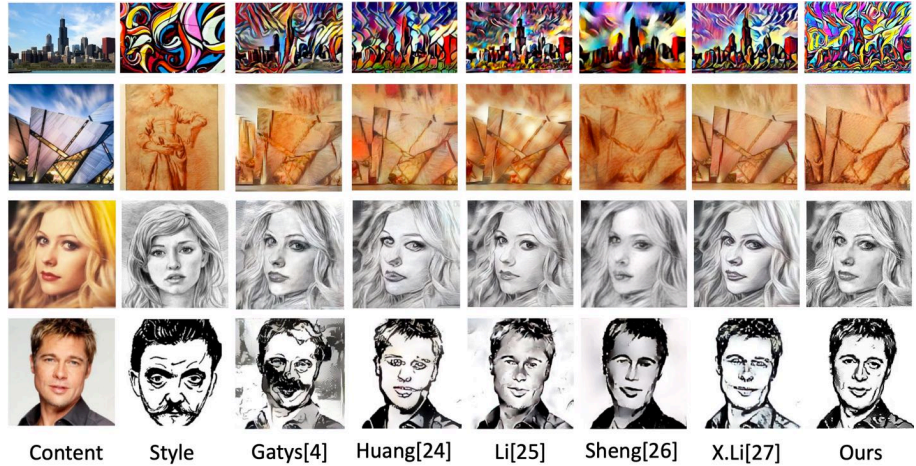

Figure 13: Comparison of artistic style transfer. These four sample pairs and five results from the third column to the seventh column are all from [27]. Our results are shown in the last column, all generated images are initialized from content image.

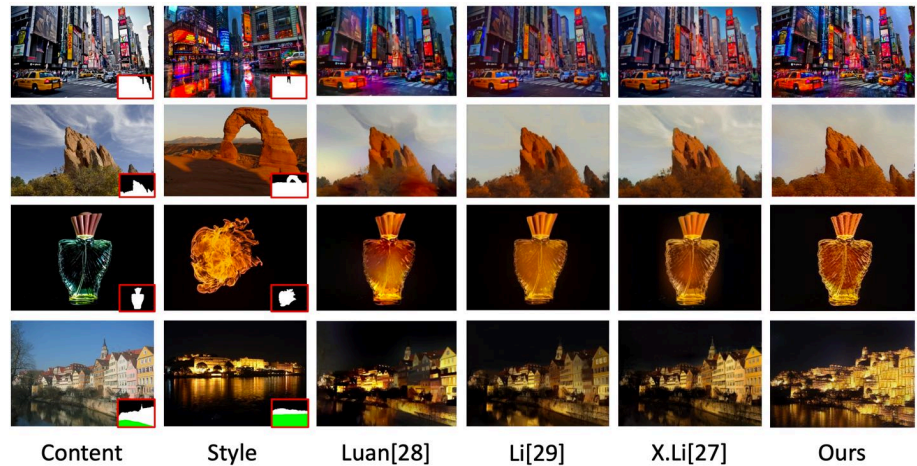

Figure 14: Comparison of photo-realistic style transfer, spatial mask is displayed at the right bottom corner of each content and style image. These four sample pairs and three results from the third column to the fifth column are all from [27]. Our results are shown in the last column, all generated images are initialized from content image.

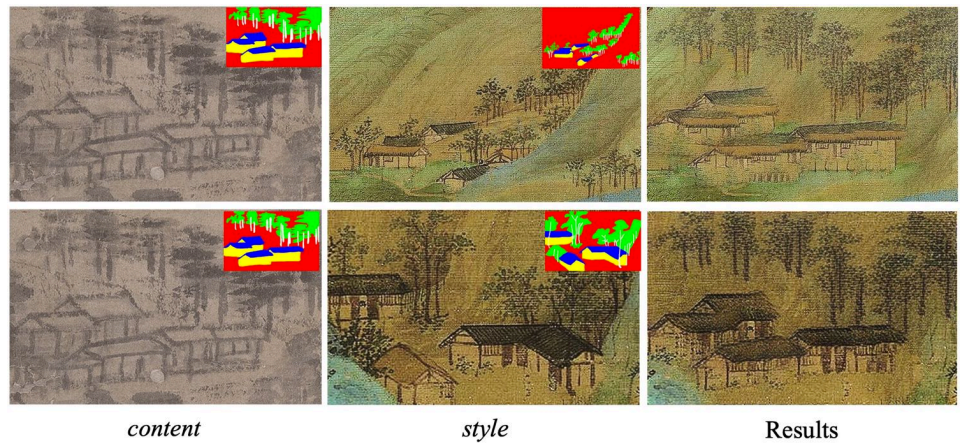

Figure 15: Detail region style transfer of Chinese ancient paintings. These regions (column 1 and column 2) are all cropped from content painting and style painting in Figure 1, respectively. We use more elaborate semantic map to get more sophisticated results. Our approach can obtain good detail performance (see last column).